# CycleGAN for Undamaged-to-Damaged Domain Translation for Structural Health Monitoring and Damage Detection

Furkan Luleci[1]; F. Necati Catbas[2*], Ph.D., P.E.; Onur Avci[3], Ph.D., P.E.

[1]Graduate Research Assistant, Department of Civil, Environmental, and Construction Engineering, University of Central Florida, Orlando, FL, 32816, USA (Email: furkanluleci@knights.ucf.edu)

[2*]Professor, Department of Civil, Environmental, and Construction Engineering, University of Central Florida, Orlando, FL, 32816, USA (Email: catbas@ucf.edu)

[3] Research Assistant Professor, Civil, Construction, and Environmental Engineering, Iowa State University, Ames, IA, 50011, USA (Email: oavci@iastate.edu)

**Abstract:** The recent advances in the data science field in the last few decades have benefitted many other fields including Structural Health Monitoring (SHM). Particularly, Artificial Intelligence (AI) such as Machine Learning (ML) and Deep Learning (DL) methods for vibration-based damage diagnostics of civil structures has been utilized extensively due to the observed high performances in learning from data. Along with diagnostics, damage prognostics is also vitally important for estimating the remaining useful life of civil structures. Currently, AI-based data-driven methods used for damage diagnostics and prognostics centered on historical data of the structures and require a substantial amount of data for prediction models. Although some of these methods are generative-based models, they are used to perform ML or DL tasks such as classification, regression, clustering, etc. after learning the distribution of the data. In this study, a variant of Generative Adversarial Networks (GAN), Cycle-Consistent Wasserstein Deep Convolutional GAN with Gradient Penalty (CycleWDCGAN-GP) model is developed to investigate the "transition of structural dynamic signature from an undamaged-to-damaged state" and "if this transition can be employed for predictive damage detection". The outcomes of this study demonstrate that the proposed model can accurately generate damaged responses from undamaged responses or vice versa. In other words, it will be possible to understand the damaged condition while the structure is still in a healthy (undamaged) condition or vice versa with the proposed methodology. This will enable a more proactive approach in overseeing the life-cycle performance as well as in predicting the remaining useful life of structures.

## 1. Introduction

Structural Health Monitoring (SHM) has been becoming more prominent for condition assessment as the civil structures continue to age over the years. Aging structures are getting vulnerable to manmade or environmental stressors, which decreases their performance and functionality and consequently shortens the remaining useful lives of the structures. This is especially true for today's world, where catastrophic events that damage structures occur frequently and are forecasted to be more. It is, therefore, essential to implement an efficient health management plan to improve the life cycle of civil structures. Consequently, it is critical to maintain structural safety, in other words, to avoid structural failures in a proactive manner, and more importantly, to save human lives.

### 1.1. Brief Review: SDD of Civil Structures

Structural Damage Diagnostics (SDD) or Damage Identification (Damage-Id) and evaluation based on the collected data from the civil structures are the main goals of SHM. The SDD procedure is a "vibration-based" method that utilizes the dynamic response of structures as in additional areas in structural engineering such as vibrations serviceability (Barrett et al. 2006; Catbas et al. 2017). The definition of damage here is any changes to the geometric and/or material properties such as cracks, delamination, spalling, corrosion, bolt loosening, etc., which negatively affect the performance and functionality of the structure. For that reason, SDD is described as a five-step process in SHM (Farrar and Worden 2007): Level-I (damage detection: indication of the existence of the damage), level-II (damage localization: indication of the location of the damage), level-III (damage classification: indication of the type of the damage), level-IV (damage assessment: indication of the severity of the damage) and level-V (damage prognosis: indication of the estimated remaining useful life of the structure). The SDD levels are linked to each other as the higher levels usually require information from the lower levels. (Avci et al. 2021) classified the structural damage assessment techniques considering level I to IV into two: (1) Local Methods such as Non-Destructive Testing (NDT) and camera sensing techniques (IR, DIC, RGB), and (2) Vibration-based Methods (Global) such as analyzing the collected vibration data parametrically (utilizing physical model, FEA software, or non-physical model like system identification techniques to obtain the structural parameter) or nonparametrically (using statistics-based methods on the raw vibration data to find the useful features then to classify or regress, or to detect anomalies). The damage assessment starts with collecting data. Data collection with accelerometers that are installed on the structure is a commonly used method to implement vibration-based SDD since it is advantageous over other methods (Catbas and Aktan 2002; Das et al. 2016). Thus, in this study, the authors focused on vibration-based SDD applications (Do et al. 2018). In literature, some of the SDD studies that do not use Machine Learning (ML) or Deep Learning (DL) methods (Gul and Catbas 2008, 2011; Krishnan Nair and Kiremidjian 2007; Silva et al. 2016; Yin et al. 2009) are utilized for SDD applications. Recently, due to their high performance in learning the complex data structures, the researchers employed ML and DL algorithms towards SDD problems successfully (Abdeljaber et al. 2017, 2018; Abdeljaber and Avci 2016; Avci et al. 2017; Bandara et al. 2014; Cury and Crémona 2012; Eren 2017; Ghiasi et al. 2016; González and Zapico 2008; Lee et al. 2005; Lee and Kim 2007; Pathirage et al. 2018; Rastin et al. 2021; Santos et al. 2016; Shang et al. 2021; Silva et al. 2016; Yu et al. 2019). More recently, the authors pointed out the data scarcity problem in SHM which hinders the use of ML and DL models for SDD applications. They demonstrated that Generative Adversarial Networks (GAN) could be efficiently used for vibration-data generation to support the DL-based SDD applications (Luleci et al. 2021a; b, 2022). The previously mentioned studies are conducted at level-I to level-IV. Level-V SDD aims to predict the Remaining Useful Life (RUL) of the structure. Level-V SDD (damage prognostics) can be categorized into three groups. (1) Model-based methods (An et al. 2013; Kwon et al. 2016; Ortiz and Carrasco 2016) estimate the RUL using the established mathematical models of the structure. Due to the size and complexity of real civil structures, it is difficult to build numerical models and create labeled data for the damaged conditions to be used for training and testing for damage detection purposes. These models are centered on the historical data which rely on the parameters of the structure and require a priori. (2) Data-driven methods (Sayyad et al. 2021; Wang et al. 2020; Wu et al. 2018) predict the RUL based on the collected operational data from the structure, and these methods are not based on the physical parameters of the model to establish a prediction model. These methods are quite favored in practice since they can be efficient and inexpensive, cheaper, and can cover a wider scope of the system. Yet, they have a larger confidence interval and require a substantial amount of data to train the model, e.g., ML or DL model. (3) Hybrid methods (Liao and Kottig 2014; Medjaher et al. 2012) use both model-based and data-driven techniques to overcome the limitations of both methods. Nevertheless, currently, the "Data-driven" and "Hybrid" methods are the prominent approaches in prognostics due to superior prediction skills of the used ML and DL models. One of the common details in the above-mentioned methods in SDD is that they all use historical data from the structure to directly form the prediction model to eventually achieve the SDD levels. In the AI field, they are named "data-hungry" models which are dependent on a substantial amount of data to establish a robust AI model. Although some of these methods are generative-based models which do not establish a decision boundary in the observed data like discriminative models, they learn how the distribution of the data domain is shaped. Yet, after learning the data, they are generally employed to do ML or DL tasks such as classification, regression, clustering, except the most recent conducted studies (Luleci et al. 2021a; b, 2022). The research gap in SDD applications is the lack of use of a robust generative-based model, which can be employed to provide possible future responses of civil structures to enable superior predictive condition assessment.

### 1.2. Motivation and Objective of the Study

The employed methodologies in the SHM field significantly rely on the collected data from civil structures. Particularly, with the advancements in the AI field, AI-based methods will be utilized more requiring a substantial amount of data to establish the prediction models. Yet, it is widely known in the SHM field that data collection from civil structures is usually a challenging and expensive task. Some of the difficulties are obtaining permission from authorities to install costly and laborious SHM systems, requesting traffic closures, setting up communication networks between sensors and central data acquisition systems. In addition, the opportunity of obtaining before and after damage-associated data from the structures is very rare. Also, considering that very few structures are deployed with permanent SHM systems, it is not possible to determine the damage state of the other structures. Some data generation solutions are briefly discussed in (Luleci et al. 2021b), such as using a Finite Element model, but this can be unreliable, timely, and inaccurate. Recently, studies presented in (Luleci et al. 2021a; b, 2022) demonstrated a methodology to tackle the data scarcity problem in SHM by providing synthetic damaged data samples for more accurate data-driven models such as AI-based methods (ML and DL algorithms). These studies point out that the presented method decreases the need for data collection from structures and solves the class imbalance problem for DL-based SDD applications. Yet, their model, Generative Adversarial Networks (GAN), is only capable of learning one domain (damaged domain) and generating similar samples for the same domain (damaged-to-damaged domain translation). The usage of that methodology might be limited to only data generation for AI-based SDD applications. **The objective of the study presented here is to translate one domain to another domain by using a GAN tool. In this study, the authors employ a GAN variant, CycleWDCGAN-GP (Cycle-Consistent Wasserstein Deep Convolutional Generative Adversarial Networks using Gradient Penalty), to learn the mapping between two different vibration data domains for generating unhealthy response data of a civil structure from existing healthy response data (or vice versa).** Producing potential future dynamic responses of a civil structure enables the implementation of both parametric (e.g., SDD with system identification algorithms) and nonparametric (e.g., SDD on raw response vibration signal) tools for the broader use of SDD. The authors of this study name this process **Undamaged-to-Damaged vibration data domain translation** which is classified into two tasks: **Undamaged to Damaged (U2D) and Damaged to Undamaged (D2U)**. The data used in this study for both domains are sets of vibration responses recorded from a grand-stand steel structure under a white noise excitation. With this methodology, it will be possible to understand potential deviation on dynamic structural response due to future damage conditions. As such, taking predictive action items after analyzing the artificially created dynamic response data will provide significant advantages over existing predictive methods for SDD applications. Additionally, utilizing this technique will address and reduce the data scarcity problem for structures since the model will be able to produce potential future dynamic response data. Thus, it would facilitate the condition assessment of civil structures extensively by enabling the analysis of possible future dynamic responses of the structures; thus, more accurate predictions on all levels of SDD could be carried out.

### 1.3. Brief Review: GAN

In 2014, Goodfellow et al. (Goodfellow et al. 2014) introduced a generative-based AI model, Generative Adversarial Networks. The model consists of a generative, $G_\theta$, and a discriminative network, $D_\varphi$. While the $G_\theta$ captures the given random noise data, $z$, and then maximize the probability that the data it generates, $G_\theta(z)$, as similar as the real ones, $x$, the $D_\varphi$ aims to learn the $x$ and make predictions on the received samples from both domains, $G_\theta(z)$ and $x$. In that regard, GAN is considered a two-player game where each player tries to trick each other as well as learn from each other. The introduced loss function to train the GAN in the paper is shown in Equation 1.

$$\min_\theta \max_\varphi V(G_\theta, D_\varphi) = \mathrm{E}_{x \sim pdata(x)}[log D_\varphi(x)] + \mathrm{E}_{z \sim pz(z)}[log(1 - D_\varphi(G_\theta(z)))] \qquad (1)$$

The authors of (Goodfellow et al. 2014) demonstrated that the GAN shows outstanding performance on image generation. As a result, many other researchers conducted GAN-based studies. Most of these studies focused on the challenges of training GAN such as the model being hard to converge due to finding a unique solution to Nash equilibrium (the balance between two networks – generator and discriminator). Results suffer from large oscillation in loss values of generator and discriminator during the training which makes the training unstable thus, reaching convergence in the model's formulation gets very difficult. Another difficulty of training GAN is "mode collapse", where the generator learns only some specific features in the given domain which can trick the discriminator with the same generated outputs during the training. As a result, the generator keeps producing the same output which reduces the diversity of the outputs. Also, intuitively at the beginning of training, the discriminator performs better since it has more knowledge about the data domain than the generator has, thus it rejects all the outputs generator produces. At this point, this could fail the training due to vanishing gradients, hence the generator does not learn from the discriminator since it cannot find the weak spot of the discriminator. Researchers introduced some tips, namely "hacks" to alleviate these drawbacks of training GAN (Goodfellow 2016; Salimans et al. 2016). Thereafter, (Radford et al. 2015) proposed using deep convolutions in training GAN after their successful adoption in the Computer Vision field. They noted that using **deep convolutions** helped the GAN training significantly better and learned the domain representation more effectively. (Arjovsky et al. 2017) addressed the difficulty of training GAN. They proposed a radical solution that uses Wasserstein distance as the cost function in GAN. Wasserstein distance is a distance function that measures the distance between probability distributions in metric space. Thus, it has a more meaningful loss value (directly related to the accuracy of the generated outputs) than the previous generative models' cost functions such as KL divergence and JS divergence. Therefore, the authors did not include the sigmoid function and the model only outputs a scalar value other than a probability. The output values of this function state how much the input data is close to real samples, thus they converted the name of the discriminator to "critic". The authors named the model **Wasserstein Generative Adversarial Networks (WGAN)** and demonstrated that it improves the training with a more stable training process and accurate outputs. To calculate the "Wasserstein distance", the 1-Lipschitz function has to be defined. To do that, the authors implemented a very simple weight clipping to restrict the weights within a certain range controlled by a parameter. However, the "weight clipping" method causes the model to lower its learning capacity and also the model becomes highly sensitive to its parameters. For instance, if the weight clipping value is selected as high, the computation time increases significantly and if it is small, the "vanishing gradients" problem occurs. To solve the weight clipping problem, (Gulrajani et al. 2017) introduced gradient penalization in the critic's loss function instead of weight clipping. This methodology demonstrated more stable performance in training and produced higher quality images. As a result, the authors named the model **Wasserstein Generative Adversarial Networks using Gradient Penalty (WGAN-GP).** Furthermore, it is proved that batch normalization negatively impacts the gradient penalty, thus instance normalization is used. In 2017, (Zhu et al. 2017) pointed out an important problem for supervised learning algorithms. In order to train supervised models, they require to have labeled pairs of corresponding data from two domains. But in reality, most of the time, labeled paired data is difficult to obtain and even not possible in some cases. This is especially true for civil SHM applications where useful data (damaged domain data) is rare to find. The authors proposed a GAN-based model where the model can capture the mappings in two domains and learn how these mappings can be translated into other domain set without needing to have a paired dataset. In this model, the authors use two GAN models where the produced output of the first generator is used by the second generator. Then, the output of the second generator should be the same as the original input that is used by the first generator. The reverse is also true. That being said, the authors used a loss function, "cycle consistent loss", as an additional loss to the adversarial losses. The cycle consistent loss acts as a regularization in the generators to encourage the model to learn the mapping in both domains. The authors named the model **Cycle-Consistent Adversarial Networks (CycleGAN).** They used CycleGAN on various image datasets and translated one image domain to another domain successfully. Perhaps the most notable one is translating horse pictures into zebra pictures by using the trained model's knowledge. Note that the image samples in both domains are not paired and randomly taken images of horses and zebras. Recently, a study (Hu et al. 2021) trained the CycleGAN with Deep Convolutions, Wasserstein loss, and Gradient Penalty methods. They found out that while applying Wasserstein loss only improved the training, applying both Wasserstein loss and Gradient Penalty, which the authors named **CycleWGAN-GP**, enhanced the image quality. Yet, they addressed that using the CycleWGAN and CycleWGAN-GP models in different applications needs to be investigated further since their study is the first instance of such.

GAN is largely adopted in the Computer Vision field on image applications (2-D data). Except for using GAN for image datasets, some other disciplines utilized the GAN successfully for different applications such as 1-D data generation, reconstruction, filtering, data conversion (Ferro et al. 2019; Gao et al. 2019; Gu et al. 2021; Guo et al. 2020; Kaneko et al. 2019; Kaneko and Kameoka 2017; Kuo et al. 2020; Li and Wang 2021; Li et al. 2021; Luo et al. 2020; Sabir et al. 2021; Shao et al. 2019; Truong and Yanushkevich 2019; Wang et al. 2021; Wulan et al. 2020; Xiang et al. 2020; Yao et al. 2018; Zhang et al. 2021). In the civil SHM field, few GAN-based studies are presented for 1-D data reconstruction and generation (Fan et al. 2021; Jiang et al. 2021; Luleci et al. 2021a; b, 2022; Zhang et al. 2018). Yet, to the best knowledge of the authors, there has not been a study using CycleGAN or its variants in the civil SHM field. Therefore, the authors believe that **the presented methodology here in this paper on CycleWDCGAN-GP will pave the way for faster, more accurate, more efficient, and less data collection need for SHM and SDD of civil infrastructure.**

2. Workflow

The study starts with introducing the dataset that was collected from a steel grandstand structure and available in benchmark dataset format (Avci et al. 2022), Step (1) of Fig 1. Then, the preprocessing of the dataset is explained followed by the building of the CycleWDCGAN-GP model and training it, Step (2) of Fig. 1. Then, the model performance indicators such as model losses, metric and statistical indices, and Fast Fourier Transform (FFT) are explained for monitoring the training and evaluation of the generated outputs of the model to the real datasets in both time and frequency domain. Subsequently, utilizing the mentioned indicators, the training process is monitored and evaluated, Step (3). After the training, the model is used to produce the damaged vibration domain by using the undamaged vibration domain (U2D) and vice versa (D2U), generating the undamaged vibration domain by using the damaged vibration domain, Step (4) of Fig. 1. Then, the model is evaluated using the aforementioned indicators to measure the performance of the used model, step (5).

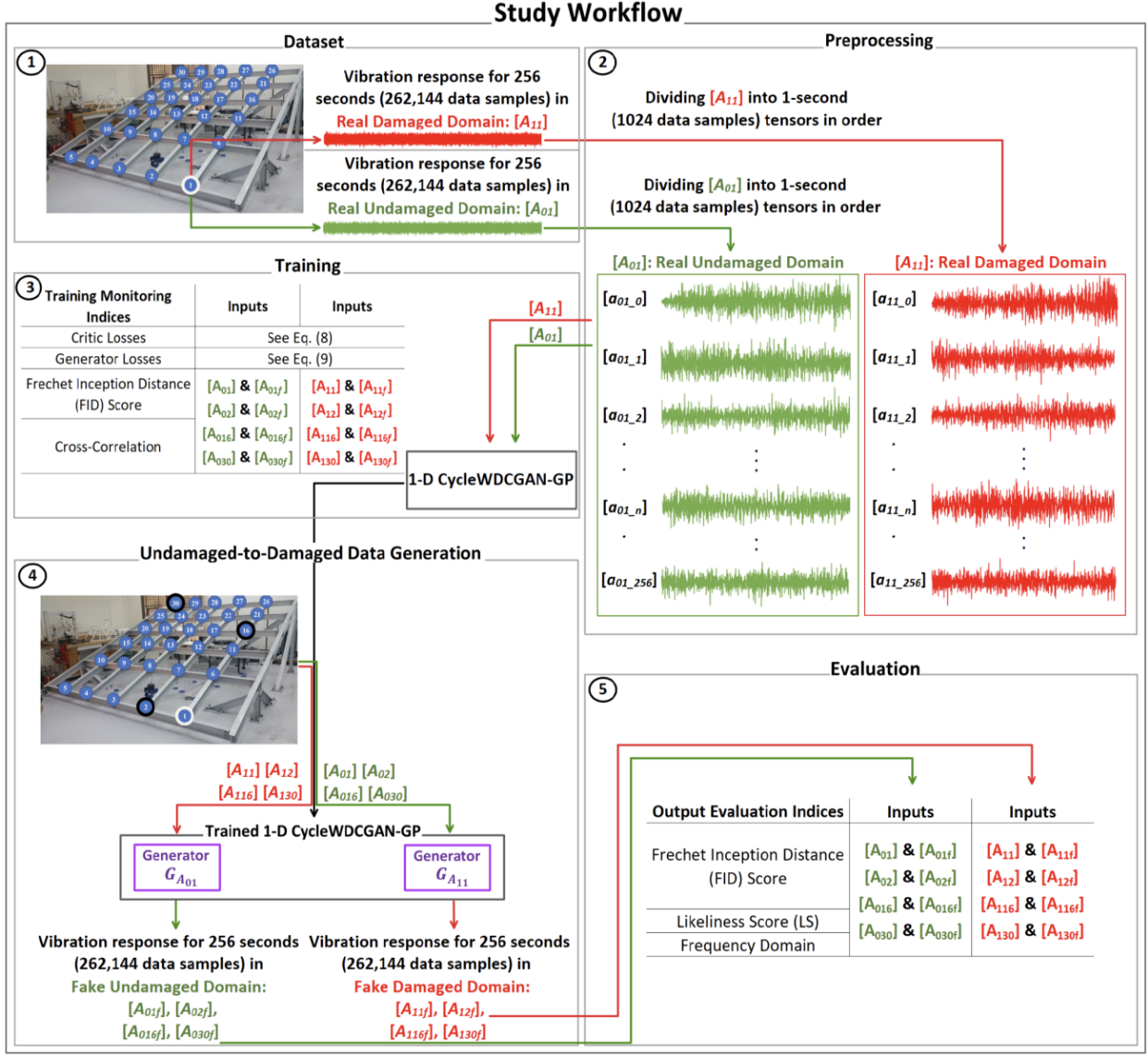

**Fig. 1.** Study Workflow

In order to explain the data flow in the paper clearly, some notations are defined for the used 1-D vibration datasets, which are named "tensors". The notation used for the tensors are $[A_{cjf}]$ and $[a_{cjf\_n}]$ where $A$ represents the entire vibration tensor (262,144 samples), $c$ is the condition of the domain whether it is undamaged or damaged, $j$ is the number of joints in which the acceleration is collected at, $f$ denotes whether the tensor is the real (ground truth) or the generated (synthetic – "fake") one, $a$ is the tensor after its division by 256 tensors (each tensor contains 1024 samples - 1 second tensor), and $n$ is the number of the vibration tensor in the entire vibration tensor.

## 2.1. Dataset

The workflow of the implemented methodology in this study follows steps (1) to (5) in order as shown in Fig. 1. The used dataset is obtained from a conducted study (Avci et al. 2017) where a total of 30 accelerometers were installed at the joints of a grandstand steel structure. After that, the damage is introduced at each joint in the form of bolt loosening. With that said, 30 different damage and 1 undamaged scenario were created corresponding to each joint. Then, a modal shaker applied a random white noise on the structure and the vibration data is collected at each joint at a sampling frequency of 1024 Hz for 256 seconds, which is a total of 262,144 samples. As shown in Fig. 1, only the datasets of joint 1, undamaged and damaged datasets, are utilized for training, and then the trained model is tested for joint 1 and for randomly picked Joints 2, 16, and 30. Since the model is trained on only joint 1's datasets, it is expected that the model generates very similar outputs to the real datasets of joint 1 (because it learns the undamaged and damaged domain of Joint 1). Thus, it is concerning that learning only the bolt loosening effect in the response data might not be sufficient to guess the response of other joints; this is because the damage not only affects the response at one particular joint but also affects different locations of the structure's response and the introduced damage at each location can have different influences in the structural response as well. With respect to that, it is observed that the frequency domain of each of these datasets (undamaged and damaged) are quite different from each other. The statistical parameters of the datasets of these joints are also investigated. While the observed mean values are almost the same, variance values show small differences. This is expected since the mean values of vibration response signals usually tend to be identical to each other at zero, but variance could show slight dissimilarity. Therefore, analysis in the frequency domain is expected to be more effective for the comparison of vibration response signals. This is discussed in more depth in the following sections.

### 2.2. Preprocessing

After obtaining the dataset, damaged domain, $A_{11}$, and undamaged domain, $A_{01}$, are divided into 1-second signals (tensors) in the order it is obtained from the dataset as shown in Fig. 1 – Step (2). The reasoning behind implementing such a division is to increase the efficiency of the model. After several trials, it is noticed that to train the model directly with a large amount of data (262,144 data samples or 256 seconds), a needed batch size to pass the data into the model to get the best result is around the batch size of 1024. Yet, such batch size is not preferable as the quality of the model degrades (lower asymptotic accuracy for data generation), e.g., it requires a lot of memory for processing and leads to more time for convergence of the algorithm since the computational speed is significantly lower. In addition, gradient updating steps can be very large or very small which causes instability during the training that leads to poorer learning of the used data domain. The common consensus in the DL field is that using a smaller batch size in model training generally results in a lesser number of iterations to converge with higher model performance. But this highly depends on the dataset as one big batch size can be small for another. This is also confirmed in our trials with different batch sizes. Thus, it is decided to divide the entire signal with 1-second tensors to have 1024 samples per batch which our model benefitted from this significantly during the training and reflected on the accuracy of the results. Lastly, contrary to the popular practice, normalization before feeding the datasets in the model is not implemented. Intuitively, the authors think that feeding raw data directly conserves the raw features in the dataset and supports the generative skills of the CycleWDCGAN-GP model for the generation of more similar outputs to the real dataset.

### 2.3. CycleWDCGAN-GP Model and Training

As mentioned in the previous sections, the used CycleGAN model in this study adopted the Wasserstein distance, Gradient Penalty, and Deep Convolutions as they are also recently utilized in (Hu et al. 2021). Apart from that study, some first-time approaches are taken that put forth some differences, which leveraged the performance of the model. For instance, instead of using the residual layers in the middle of the generator, which is the technique used in CycleGAN models, 2 residual layers are used together after every convolution and transpose-convolution layers. Secondly, the model used by the study has a total of 28 layers, and in the literature, it is mostly 10-14. Third, batch normalization is used in the residual layer which the common practice is using instance normalization. Fourth, identity loss is used in the generator loss which is not a preferred loss in most CycleGAN models for images but is occasionally used just to preserve the same color. The model parameters, model architecture, and model data flow diagram are given in Table 1, Fig. 2, and Fig. 3, respectively.

As discussed in section 1.3 about GAN, the 1-D CycleWDCGAN-GP model employed in this study works in the same fashion as CycleGAN do. It consists of two GAN models. Each GAN operates on **Adversarial Critic Losses**, Eq.(2-3), and **Adversarial Generator Losses**, Eq.(4-5), where critic losses state the value of how much close the output of the generator to the real ones is and generator losses state the value of critic's score for each output it produces. Additionally, the two GANs are connected with **Cycle Consistency Loss**, Eq. (6), to ensure the generated data in the damaged domain, which is translated from undamaged to damaged, is as close as possible to the input of the other generator, that is to be translated from damaged to undamaged. In other words, translation of one domain from another domain and translating it back again should arrive at the same point where it is started. The other way is also true, which is generating damaged domain from undamaged domain has to be as close as possible to the input of the other generator, that is to be translated from undamaged to damaged.

**Table 1.** Model Parameters

| Symbol | Description |
| --- | --- |
| Batch Size (N) | 1 |
| Number of Epochs | 1000 |
| Learning Rate for Generators ($G_{A_{11}}$ and $G_{A_{01}}$) | $1 \times 10^{-4}$ |
| Learning Rate for Critics ($C_{A_{11}}$ and $C_{A_{01}}$) | $2 \times 10^{-4}$ |
| Critic Iterations | 20 |
| $\lambda_{Id}$ (Identity Loss) | 10 |
| $\lambda_{Cyc}$ (Cycle Loss) | 10 |
| $\lambda_{GP}$ (Gradient Penalty) | 10 |
| Optimizer (Generator, Critic) | AdamW |

Lastly, **Identity Loss**, Eq. (7), ensures that the generator should give the same output once it receives an input from a domain which actually the generator produces for. Thus, the generator should not, in fact, change the input extensively. In other words, a generator that is responsible for translating undamaged to a damaged domain, if it receives damaged domain as an input to translate it into the damaged domain again, there should not be any changes on the output since it is already damaged when it was received by the generator. The entire model is trained based on the minimization of both **Total Critic Losses**, Eq.(8), and **Total Generator Losses**, Eq.(9), by using the Adam algorithm with weight decay regularization (AdamW) (Loshchilov and Hutter 2017). Furthermore, the lambda parameters for critic, cyclic, and identity losses are introduced in their original papers to allow the users to configure the loss functions during fine-tuning. Lastly, for gradient penalty, L2 norm Eq. (2-3), and for identity and cycle losses, Eq. (6-7) L1 norm is used. Fig. 3 can be followed with the shown equations below, Eq. (2-9).

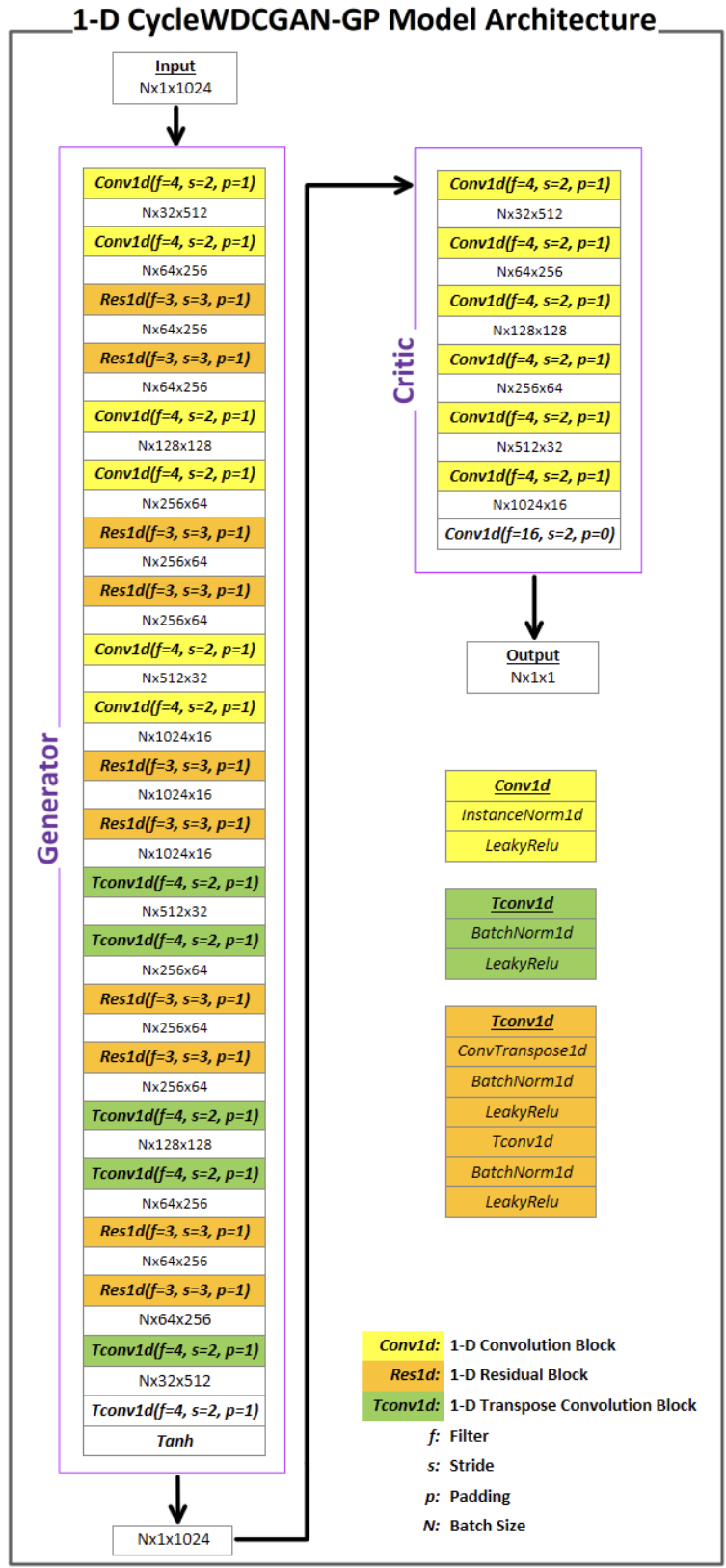

**Fig. 2.** The model architecture of 1-D CycleWDCGAN-GP

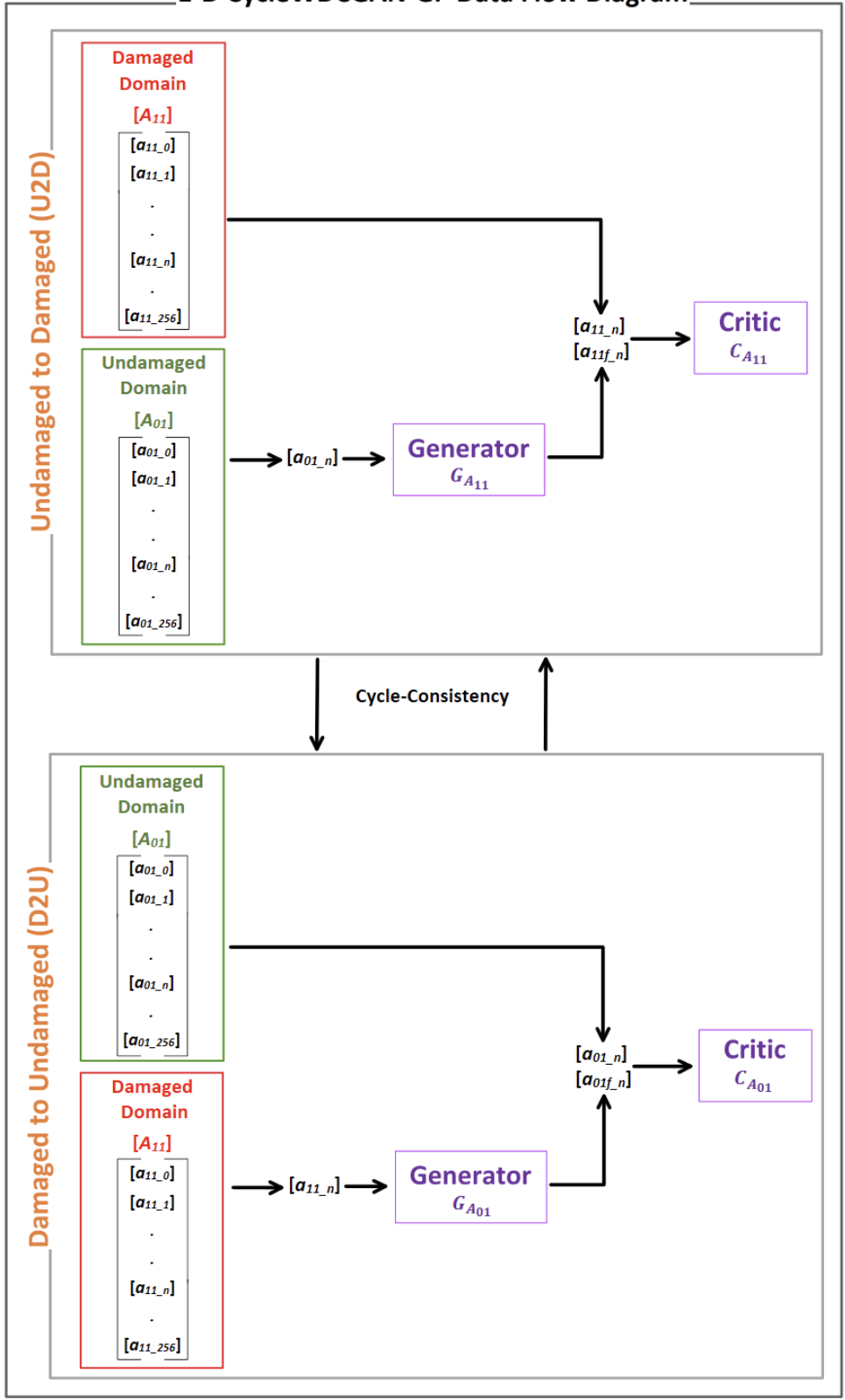

**Fig. 3.** Data flow diagram of 1-D CycleWDCGAN-GP

$$Adversarial\ Critic\ Loss_{C_{A_{11}}} = \mathcal{L}_{WDCGAN-GP}^{C_{A_{11}}}(G_{A_{11}}, C_{A_{11}}, a_{11\_n}, a_{01\_n}, a_{11f\_n}) =$$
$$\mathbb{E}_{a_{11\_n} \sim p(a_{11\_n})}[C_{A_{11}}(a_{11\_n})] - \mathbb{E}_{a_{01\_n} \sim p(a_{01\_n})}[C_{A_{11}}(G_{A_{11}}(a_{01\_n}))] +$$
$$\lambda_{GP} \mathbb{E}_{a_{11f\_n} \sim p(a_{11f\_n})}\left[\left(\left\|\nabla_{a_{11f\_n}} C_{A_{11}}(a_{11f\_n})\right\|_2 - 1\right)^2\right] \quad (2)$$

$$Adversarial\ Critic\ Loss_{C_{A_{01}}} = \mathcal{L}_{WDCGAN-GP}^{C_{A_{01}}}(G_{A_{01}}, C_{A_{01}}, a_{01\_n}, a_{11\_n}, a_{01f\_n}) =$$
$$\mathbb{E}_{a_{01\_n} \sim p(a_{01\_n})}[C_{A_{01}}(a_{01\_n})] - \mathbb{E}_{a_{11\_n} \sim p(a_{11\_n})}[C_{A_{01}}(G_{A_{01}}(a_{11\_n}))] +$$
$$\lambda_{GP} \mathbb{E}_{a_{01f\_n} \sim p(a_{01f\_n})}\left[\left(\left\|\nabla_{a_{01f\_n}} C_{A_{01}}(a_{01f\_n})\right\|_2 - 1\right)^2\right] \quad (3)$$

$$Adversarial\ Generator\ Loss_{G_{A_{11}}} = \mathcal{L}_{WDCGAN-GP}^{G_{A_{11}}}(G_{A_{11}}, C_{A_{11}}, a_{11\_n}, a_{01\_n}) =$$
$$-\mathbb{E}_{a_{01\_n} \sim p(a_{01\_n})}[C_{A_{11}}(G_{A_{11}}(a_{01\_n}))] \quad (4)$$

$$Adversarial\ Generator\ Loss_{G_{A_{01}}} = \mathcal{L}_{WDCGAN-GP}^{G_{A_{01}}}(G_{A_{01}}, C_{A_{01}}, a_{01\_n}, a_{11\_n}) =$$
$$-\mathbb{E}_{a_{11\_n} \sim p(a_{11\_n})}[C_{A_{01}}(G_{A_{01}}(a_{11\_n}))] \quad (5)$$

$$Cycle\ Consistency\ Loss = \mathcal{L}_{Cyc}(G_{A_{11}}, G_{A_{01}}) = \left[\mathbb{E}_{a_{01\_n} \sim p(a_{01\_n})}[G_{A_{01}}(G_{A_{11}}(a_{01\_n})) - a_{01\_n}] +\right.$$
$$\left.\mathbb{E}_{a_{11\_n} \sim p(a_{11\_n})}[G_{A_{11}}(G_{A_{01}}(a_{11\_n})) - a_{11\_n}]\right] \lambda_{Cyc} \quad (6)$$

$$Identity\ Loss = \mathcal{L}_{Id}(G_{A_{11}}, G_{A_{01}}) = \left[\mathbb{E}_{a_{01\_n} \sim p(a_{01\_n})}[G_{A_{01}}(a_{01\_n}) - a_{01\_n}] +\right.$$
$$\left.\mathbb{E}_{a_{11\_n} \sim p(a_{11\_n})}[G_{A_{11}}(a_{11\_n}) - a_{11\_n}]\right] \lambda_{Id} \quad (7)$$

**Total Critic Losses** $= \mathcal{L}_{WDCGAN-GP}^{C_{A_{11}}}(G_{A_{11}}, C_{A_{11}}, a_{11n}, a_{01\_n}, a_{11f\_n}) +$
$\mathcal{L}_{WDCGAN-GP}^{C_{A_{01}}}(G_{A_{01}}, C_{A_{01}}, a_{01\_n}, a_{11\_n}, a_{01f\_n}) \quad (8)$

**Total Generator Losses** $= \mathcal{L}_{WDCGAN-GP}^{G_{A_{11}}}(G_{A_{11}}, C_{A_{11}}, a_{11\_n}, a_{01\_n}) +$
$\mathcal{L}_{WDCGAN-GP}^{G_{A_{01}}}(G_{A_{01}}, C_{A_{01}}, a_{01\_n}, a_{11\_n}) + \mathcal{L}_{Cyc}(G_{A_{11}}, G_{A_{01}}) + \mathcal{L}_{Id}(G_{A_{11}}, G_{A_{01}}) \quad (9)$

### 2.4. Performance Indicators for CycleWDCGAN-GP

Evaluating the performance of GAN can be categorized into two: qualitative evaluation and quantitative evaluation. The commonly used form of evaluation technique is the qualitative approach in which a comparison of generator's output with the real datasets. Yet, this method suffers from some limitations such as a limited amount of produced data can be viewed at a time by a human observer, and also subjectivity is introduced in the evaluation. Particularly, implementing this approach on 1-D data (signals) is not as effective as doing it on 2-D data (images). Several quantitative evaluation indices are introduced for evaluating the performance of GAN models (Borji 2018, 2021). As mentioned in the previous sections, for nonparametric SDD, it is critical that the real and synthetic datasets are similar to each other in the time domain. For parametric SDD, the real and synthetic datasets' frequency values have to match so that it can be concluded that they are a similar form of vibration signals. For time-domain analysis, FID score and LS are used. For the frequency domain analysis, X-cross, and FFT are used which are explained below. Note that the FID, LS, X-cross, and FFT are calculated between the entire (262,144 samples) real and synthetic ("fake") vibration signals such as X-cross of $[A_{01}]$ and $[A_{01f}]$.

**Fréchet Inception Distance (FID)** (Heusel et al. 2017) is the most preferred score to evaluate the outputs of GAN since it proved its effectiveness against other methods. The equation of FID is given below in Eq.(10).

$$FID(x, g) = ||\mu_x - \mu_g||_2^2 + Tr(C_x + C_g - 2(C_x C_g)^{0.5}) \qquad (10)$$

The $\mu_x$ and $\mu_y$ are the means, $C_x$ and $C_g$ are the covariance matrices for real and generates outputs, respectively, and $Tr$ is the trace of the matrices such as summation of the diagonal elements in the matrices. The lower the FID score, the more similar the generated and the real datasets. Hence, the FID score is one of the indicators that is used by this study. **Cross-correlation (X-cross)** is another indicator that states the similarity of two signal datasets. The computation can happen in two ways, in the time domain and the frequency domain. In this study, cross-correlation is carried out in the frequency domain to monitor the similarity of the, which its formula is given below in Eq.(11).

$$x(t) \star y(t) = \mathcal{F}^{-1}(\mathcal{F}\{x(t)\}(\mathcal{F}\{y(t)\})^*) \qquad (11)$$

The $x(t)$ and $y(t)$ represents two signals to be cross-correlated and $\star$ is the cross-correlation operator. $\mathcal{F}$ and $\mathcal{F}^{-1}$ denotes the Fourier and inverse Fourier transform respectively, and the asterisk * denotes the complex conjugate. The higher the similarity between the signals, the higher value the cross-correlation gives. Cross-correlation is a highly used method to detect the resemblance between each pair, yet this has not been attempted in literature for the evaluation of GAN models. In this study, a strong relationship is found between the cross-correlation of real and synthetic datasets. Another indicator that is used by this study is the **Likeliness Score (LS)** (Guan and Loew 2020), which is a novel indicator to evaluate the performance of GAN based on the separability of the real and synthetic datasets. When the separability is high, the two domains have high differences. Thus, the authors introduced the Distance-based Separability Index (DSI), and accordingly, they suggested the Likeliness Score (LS) by subtracting from the value of one. In this regard, high LS indicates a high probability of two domains being similar and low LS (when the domains are not similar). The highest LS can be 1, and the lowest can be 0. For more information about the formulation of DSI, readers are referred to (Guan and Loew 2020). Lastly, the **Fast Fourier Transform (FFT)** is utilized to examine the real and synthetic datasets in the frequency domain.

### 2.5. Monitoring the Training of CycleWDCGAN-GP

The monitored indices during the training are Critic Loss, Generator Loss, FID scores and Cross-Correlation of the real and generated tensors. From Fig. 4 below, it is seen that critic loss started converging at zero around epoch 400. While the generator, at first, seems to be getting converged at epoch 400 as well, after a small upwards move, it tends to go down slightly after epoch 540. The full convergence of the generator is anticipated after a few hundred epochs. The critic and generator loss graphs show fine correlation with cross-correlation graphs as they follow similar trendlines in the plots. The FID scores are seemed to be converging first with a sharp decline in the plots before other indicators (for a more in-depth understanding of FID scores for different models, readers are referred to the other GAN studies (Costa et al. 2019; MITCSAIL 2019)). However, for vibration signals, the variance is more important as it has more influence on the FID score since vibration signals' mean values tend to be very close to zero. Thus, the impacting factor in the FID score is actually the changes in variance; and it seems that the variance values of both domains are getting similar to each other at very early epochs. The training is stopped at epoch 1000 because it is thought the convergence is reached since the model was having very small changes in the indicators such as the loss functions, X-cross, and FID scores were not going down any further. In other words, the model was having very small updates in its gradients and taking relatively very small steps on its learning trend. It is anticipated that, if the training had continued a few hundred epochs more, the generator would have full convergence at zero. But this would have taken an even more amount of time such that this training only took 4 days 8 hours 43 minutes on PC specs of RTX 3070, 16 GB RAM. As a result, the performance indicators during the training of the model yielded satisfactory results and the loss values had convergence at zero.

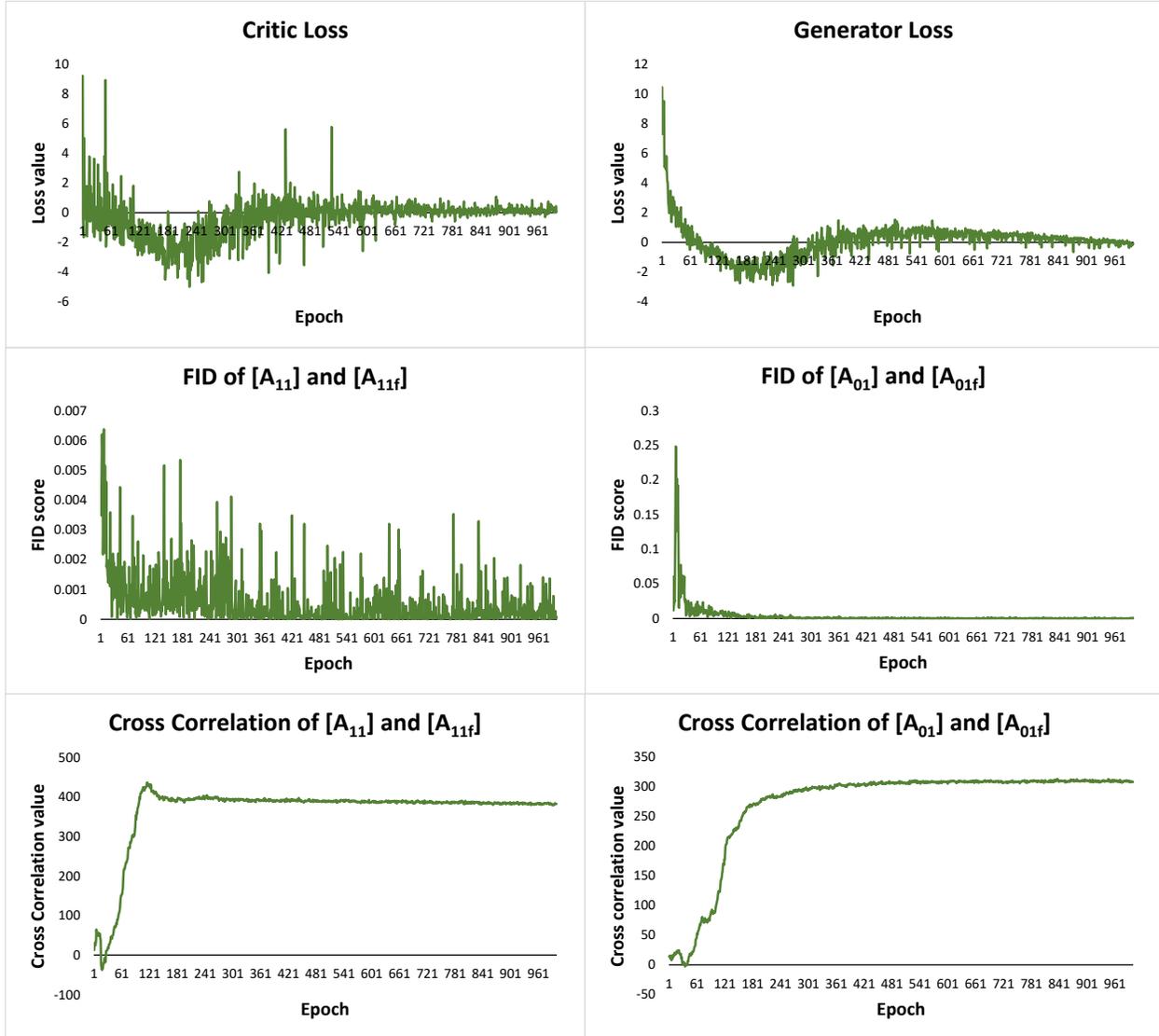

**Fig. 4.** Training monitoring plots

## 2.6. Evaluation of the Trained CycleWDCGAN-GP

After the training of the 1-D CycleWDCGAN-GP model, the U2D process is executed, which is using the trained model to translate the $[A_{01}]$ to $[A_{11f}]$, and similarly, the D2U process is executed, which is translating the $[A_{11}]$ to $[A_{01f}]$ as shown in Step (4) of Fig. 1. The same implementation is also carried out for the other mentioned Joints 2, 16, and 30. Next, FID scores and LS are computed which are given in Table 2. It is observed that both FID values are very small for Joint 1. Especially, the sharp decline of the FID values during the training and converging close to zero implies that the outputs are very similar. Yet,

FID considers the signals based on its statistical properties in the time domain, thus there is no direct link to the frequency domain of the signal which can be observed in Fig. 4., such that while the FID values are in sharp decline, the frequency-based X-cross values are seemed to have very small fluctuations. The FID values of other joints are considerably small as well, yet higher than Joint 1 which is expected since the model is trained with Joint 1's datasets. The LS values are seemed to be very close to 1.00. According to its original study (Guan and Loew 2020) and based on how close the values are to 1.00, it can be concluded that the two domains (real and synthetic domains) are very similar to each other to the extent that they cannot be separated. The LS values of the other joints are also looking relatively satisfactory, but more study is needed on this score as to what minimum LS value would be sufficient for evaluating the model. Another point that is noticed is FID and LS values follow a negative correlation as when the LS values are low, the FID values are high or vice versa. This makes sense since they are derived from similar statistical formulations. Next, the FFT of the $[A_{01}]$ and $[A_{01f}]$, and $[A_{11}]$ and $[A_{01f}]$ are computed respectively, and their power spectrums are plotted in Fig. 5. Additionally, similar implementation is carried out for the other Joints 2, 16, and 30. It can be observed from Fig. 5 that the real undamaged $[A_{01}]$ and synthetic ("fake") undamaged $[A_{01f}]$ datasets are very similar to each other in their frequency domain. Likewise, the real damaged $[A_{11}]$ and synthetic ("fake") undamaged $[A_{11f}]$ datasets are also very similar to each other in their frequency domain. This is anticipated as the model is trained on Joint 1's datasets. This similarity in frequency domain also shows great consistency with the FID and LS values of the $[A_{01}]$ - $[A_{01f}]$, and $[A_{11}]$ - $[A_{11f}]$. Yet, the same thing cannot be said for the other joints, $[A_{02}]$ - $[A_{02f}]$, $[A_{12}]$ - $[A_{12f}]$, $[A_{016}]$ - $[A_{016f}]$, $[A_{116}]$ - $[A_{16f}]$, $[A_{030}]$ - $[A_{030f}]$, and $[A_{130}]$ - $[A_{130f}]$ as their FID and LS values are as good (slightly lower than Joint 1), but their frequency domains are not closely similar. This might imply that at some point in FID and LS values range, the frequency domains start differing from each other which needs to be investigated as a future study. Nevertheless, although they are used for evaluation in the time domain, they show reasonable consistency with the frequency domain up to some point.

Fig. 6 – 8 show that the trained model on Joint 1's undamaged and damaged datasets could not really predict the undamaged and damaged responses of the other Joints 2, 16, and 30 accurately. Although a good accuracy for the generated outputs could not be obtained, the model's generated responses are not the same for each undamaged and damaged dynamic response of the joints which can be interpreted as a good generalization capability of the model (did not overfit the training data – mode collapse). Furthermore, since the model is only trained with datasets of Joint 1, it was expected that the model would not be capable of generating the undamaged and damaged outputs for Joints 2, 16, 30, which have different frequency domains than the datasets of Joint 1. This is because the 1-D CycleWDCGAN-GP model learned its weights based on the undamaged and damaged domain mappings of Joint 1. The differences in the frequency domains of the Joints 1, 2, 16, 30 can be seen in Fig. 5 – 8. Additionally, the introduced damage herein (bolt loosening) has different effects in each collected vibration response dataset. In other words, in civil structures, the frequency domains of the collected response data can vary for different locations of the structures, and on top of that different types of damages (bolt loosening, crack, delamination, stalling, corrosion) have different influences in the vibration response data. Also, when predicting the responses for different civil structures, these differences get multiplied, which gives a great number of combinations of unsimilar frequency domains. Furthermore, even the frequency domains are similar to some extent, the corresponding mode shapes of the structure to its peak frequency values might result in differences which also indicate damage in the structure. Nevertheless, the model learned the damage domain of the bolt loosening at Joint 1 and constructed a new undamaged/damaged vibration response successfully (based on its learned gradients from the Joint 1's datasets that are used to train the model) when a damaged/undamaged vibration response is given.

**Table 2.** FID and LS indicators

| Indicator | Inputs | Value |
|---|---|---|
| Fréchet Inception Distance | $[A_{01}]$ and $[A_{01f}]$ | $7.88 \times 10^{-6}$ |
| | $[A_{11}]$ and $[A_{11f}]$ | $1.76 \times 10^{-5}$ |
| | $[A_{02}]$ and $[A_{02f}]$ | $3.04 \times 10^{-3}$ |
| | $[A_{12}]$ and $[A_{12f}]$ | $3.75 \times 10^{-3}$ |
| | $[A_{016}]$ and $[A_{016f}]$ | $5.19 \times 10^{-4}$ |
| | $[A_{116}]$ and $[A_{116f}]$ | $8.25 \times 10^{-4}$ |
| | $[A_{030}]$ and $[A_{030f}]$ | $1.52 \times 10^{-3}$ |
| | $[A_{130}]$ and $[A_{130f}]$ | $1.16 \times 10^{-3}$ |
| Likeliness Score | $[A_{01}]$ and $[A_{01f}]$ | 0.99 |
| | $[A_{11}]$ and $[A_{11f}]$ | 0.96 |
| | $[A_{02}]$ and $[A_{02f}]$ | 0.80 |
| | $[A_{12}]$ and $[A_{12f}]$ | 0.80 |
| | $[A_{016}]$ and $[A_{016f}]$ | 0.86 |
| | $[A_{116}]$ and $[A_{116f}]$ | 0.85 |
| | $[A_{030}]$ and $[A_{030f}]$ | 0.83 |
| | $[A_{130}]$ and $[A_{130f}]$ | 0.82 |

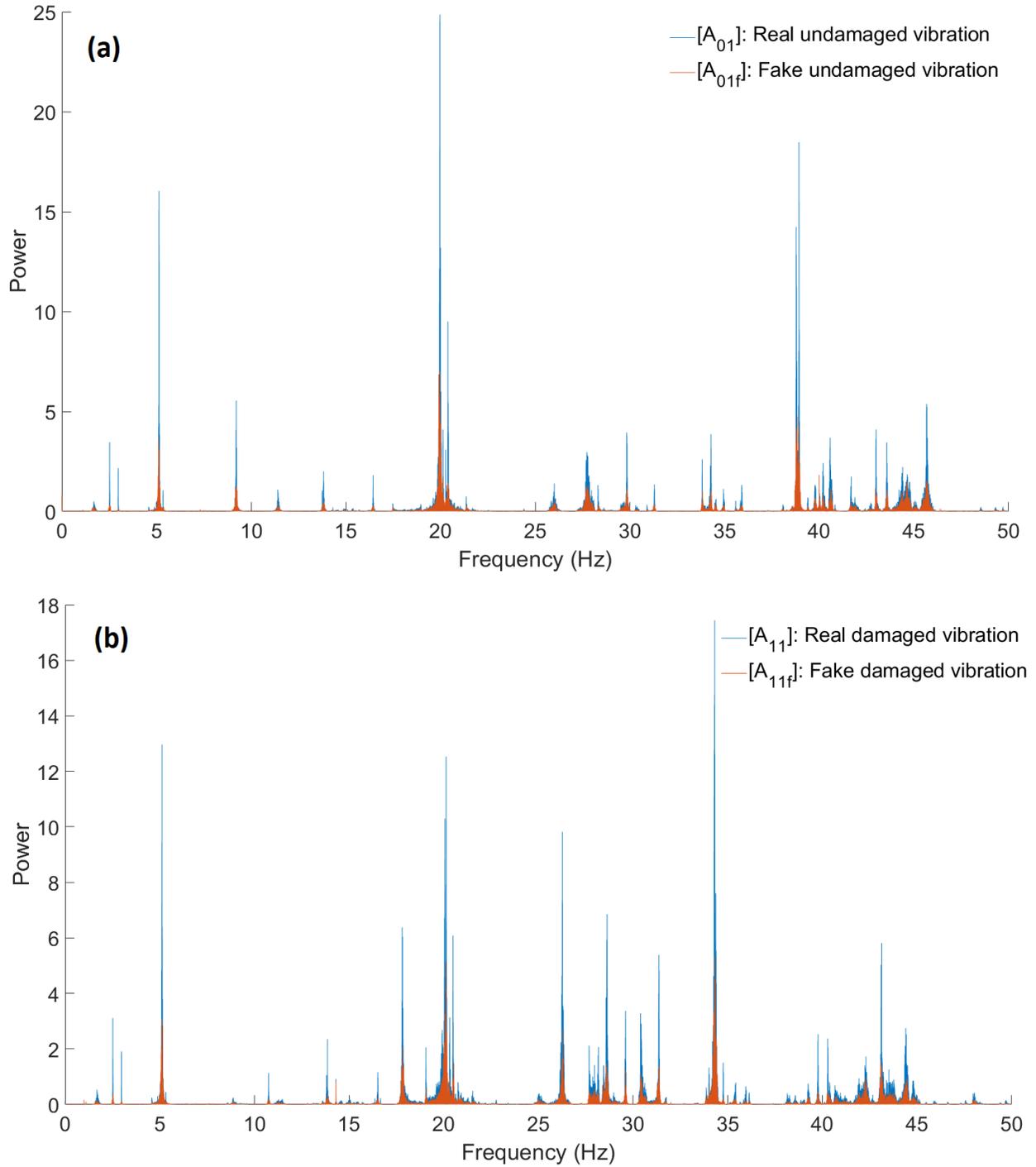

**Fig. 5.** (a) Frequency domains of real vibration tensor in the undamaged domain, $[A_{01}]$, and synthetic (fake) vibration tensor in undamaged domain $[A_{01f}]$, (b) Frequency domain of real vibration tensor in the damaged domain, $[A_{11}]$, and synthetic (fake) vibration tensor in damaged domain $[A_{11f}]$

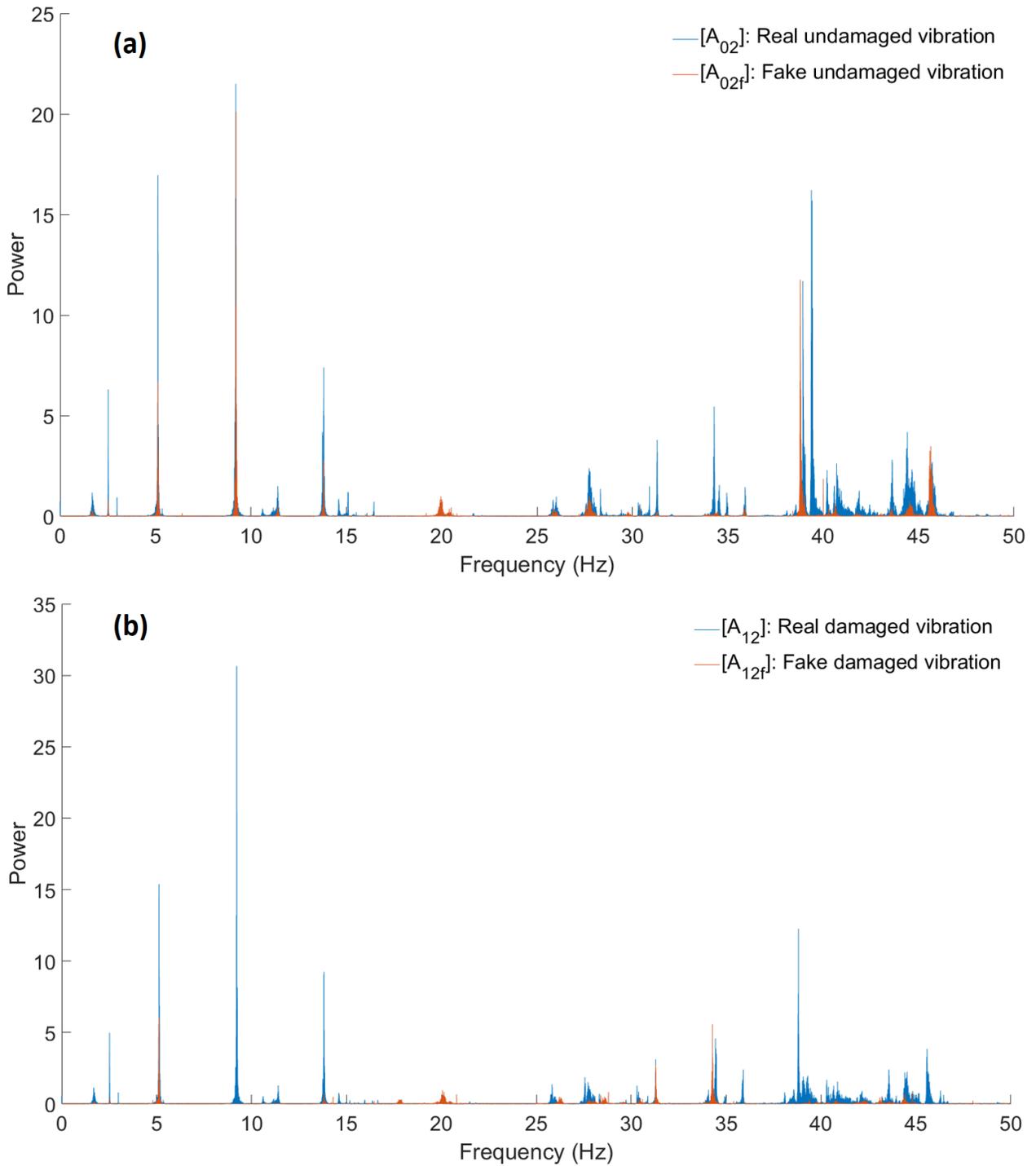

**Fig. 6.** (a) Frequency domains of real vibration tensor in the undamaged domain, $[A_{02}]$, and synthetic (fake) vibration tensor in undamaged domain $[A_{02f}]$, (b) Frequency domain of real vibration tensor in the damaged domain, $[A_{12}]$, and synthetic (fake) vibration tensor in damaged domain $[A_{12f}]$

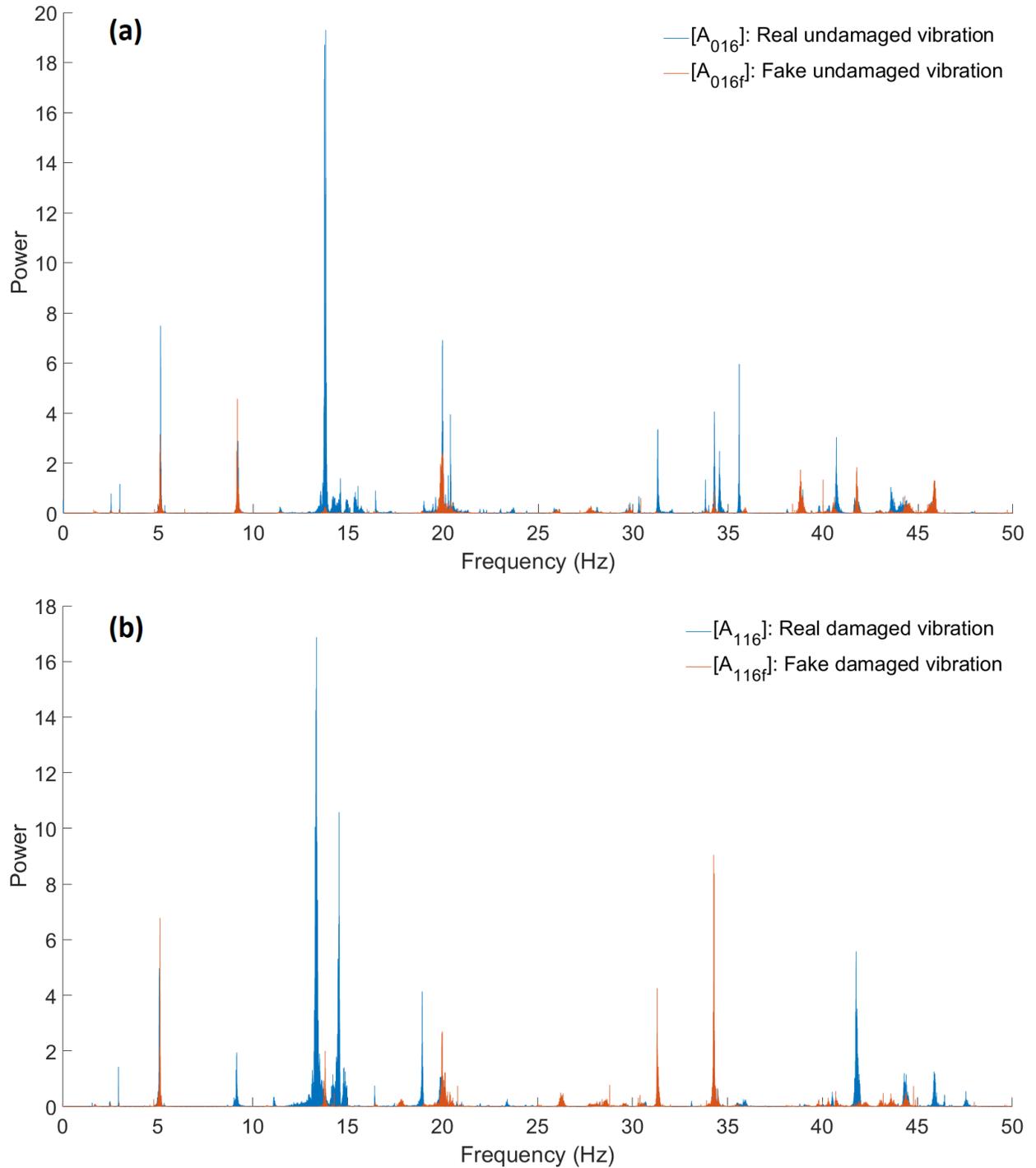

**Fig. 7.** (a) Frequency domains of real vibration tensor in the undamaged domain, $[A_{016}]$, and synthetic (fake) vibration tensor in undamaged domain $[A_{016f}]$, (b) Frequency domain of real vibration tensor in the damaged domain, $[A_{116}]$, and synthetic (fake) vibration tensor in damaged domain $[A_{116f}]$

**Joint 30**

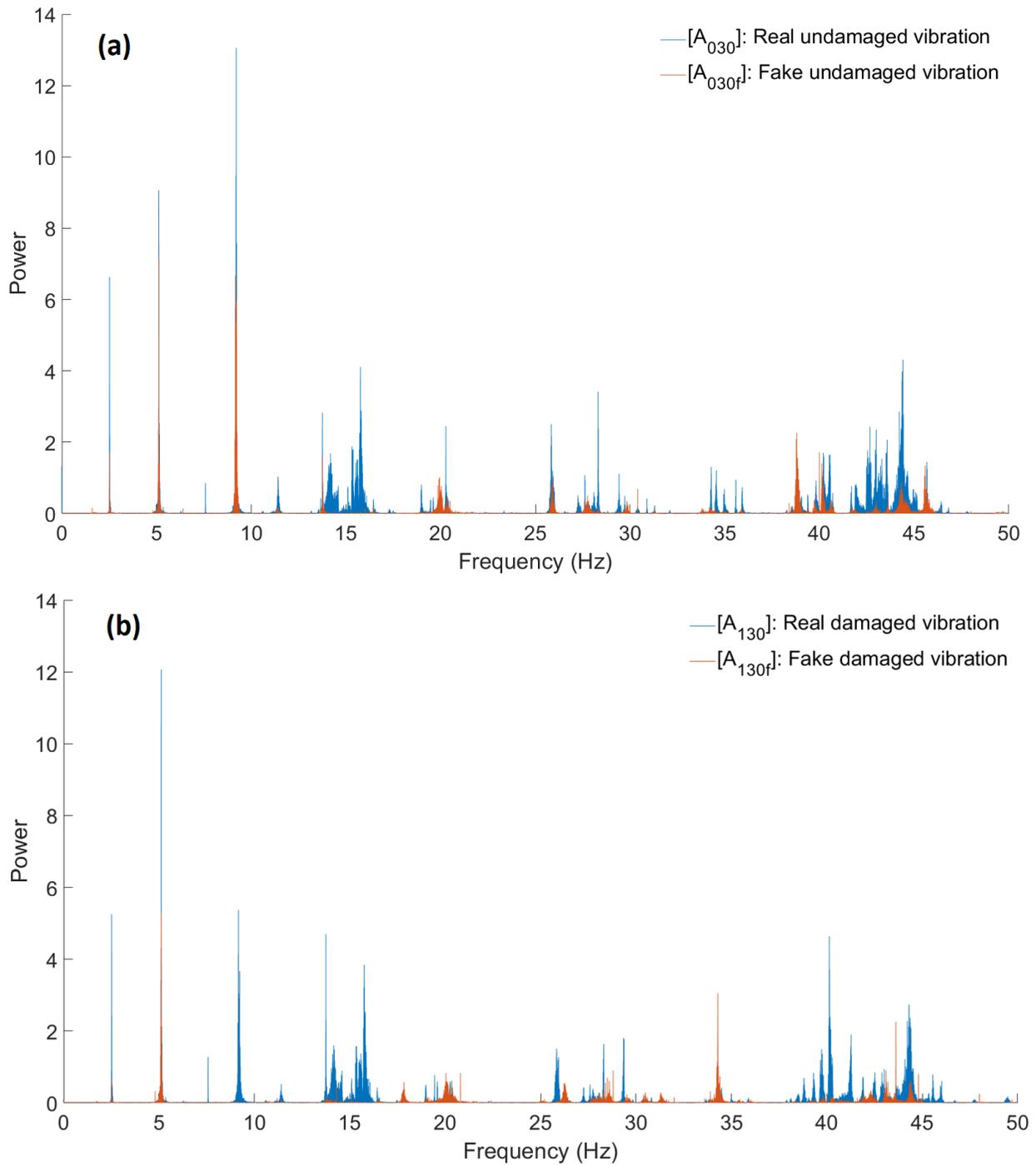

Fig. 8. (a) Frequency domains of real vibration tensor in the undamaged domain, $[A_{030}]$, and synthetic (fake) vibration tensor in undamaged domain $[A_{030f}]$, (b) Frequency domain of real vibration tensor in the damaged domain, $[A_{130}]$, and synthetic (fake) vibration tensor in damaged domain $[A_{130f}]$

## 3. Summary and Conclusions

In the last couple of decades, advances in data science have drastically influenced several disciplines and laid a solid foundation for AI applications to build on. A variety of disciplines benefited from the effectiveness of AI methods through the ML and DL algorithms. In the civil SHM field, particularly towards the SDD applications, ML and DL algorithms have been proven to be successful. Yet, ML and DL methods currently used are mostly centered on the historical data collected from the structure (labeled data) to build the prediction models to perform classification, regression, and clustering tasks to achieve acceptable SDD performance. Such methods are based on discriminative approaches, and they lack generative skills. Furthermore, data collection from civil structures is not easy, and obtaining useful data from damaged structures (damaged domain) is especially challenging. In other words, SDD applications have not exploited the opportunity to use generative-based models for superior predictive condition assessment.

Recently, Generative Adversarial Networks (GAN) and their variants impressed the scientific community with the generative capabilities on the image data applications. In this study presented herein, a variant of the GAN model, 1-D CycleWDCGAN-GP, is employed not only to learn a single data domain and generate similar data samples but also to learn the mapping between two data domains (undamaged and damaged) such that when a healthy response from a structure is input in the model, the model can translate it into an unhealthy response. In that regard, a collection of damaged (joint bolt-loosening) and undamaged dynamic response datasets from a steel frame structure is used to train the 1-D CycleWDCGAN-GP model. Subsequently, the dataset from the undamaged domain is used for translation into the damaged domain and vice versa. The performance indicators used (generator and critic losses, FID scores, LS, X-cross, FFT) to evaluate the training phase and data generation phase to confirm that the generated (translated) outputs in both domains are very similar to the "ground truth" domains (real domains). As a result, the CycleWDCGAN-GP model successfully translated the undamaged/damaged vibration domain to the damaged/undamaged domain. The model, which was trained using a particular joint data, was able to make domain translations with limited accuracy for the other particularly faraway joints of the structure, yet somewhat better results for nearby joints. A model that is trained with vibration responses collected for diverse damage types for more locations of the structure, different combinations of responses can be generated more accurately. This enables to perform vibration-based SDD with minimal data collection from structures in a very efficient way. In conclusion, this study demonstrates and confirms that GAN can learn the mapping of two different domains of vibration datasets from civil structures. In other words, the damage present in the structure can be learned by GAN, and accordingly, it would have the capacity to generate the undamaged or damaged response data from given damaged or undamaged response data of the structure. The following are some specific observations and conclusions related to this study.

- It is confirmed by this study that translation of undamaged-to-damaged vibration signal domain is possible for civil structures. The model learns the undamaged and damaged domains of a single joint and its influence in the response data and can make accurate domain translations for the same joint. Collecting large amount of data can be significantly exhaustive; however present study shows great promise to reduce this need, and the authors recommend that future studies are needed to investigate "to what extent can GAN be employed to accurately generate responses for different elements in civil structures?"

- For 1-D GAN-based studies, it is still challenging to analogize the real and synthetic datasets. The nonparametric and parametric vibration based SDD applications respectively require time domain and frequency domain evaluations. For the time domain, the currently available indicators such as the used indices in this study, FID and LS are based on statistical formulations that consider the variance and the mean values of the signals. While the variances could show slight differences, the mean values of the collected vibration signals from civil structures are identical at zero. This causes the model to be ineffective for signal comparison. Furthermore, examination of the frequency domain is needed for parametric SDD applications, yet such comparison might not be enough for SDD. Although the frequency domains of the signals suggest similarity for the signal pairs, their respective mode shapes need to be examined in detail as well in addition to frequency domain comparison. Lastly, frequency-based cross-correlation of the signal pairs is found very effective for similarity assessment of the pairs during the training as shown in this study.

**Nomenclature**

| Symbol | Description |
|---|---|
| $[A_{01}]$ | 256 seconds (262,144 samples) of vibration tensor at joint 1 in undamaged domain (0) |
| $[A_{11}]$ | 256 seconds (262,144 samples) of vibration tensor at joint 1 in damaged domain (1) |
| $[A_{01f}]$ | 256 seconds (262,144 samples) of fake vibration tensor at joint 1 in undamaged domain (0) |
| $[A_{11f}]$ | 256 seconds (262,144 samples) of fake vibration tensor at joint 1 in damaged domain (1) |
| $[a_{01\_n}]$ | $n^{th}$ 1-second (1024 samples) of vibration tensor at joint 1 in undamaged domain (0) |
| $[a_{11\_n}]$ | $n^{th}$ 1-second (1024 samples) of vibration tensor at joint 1 in damaged domain (1) |
| $[a_{01f\_n}]$ | $n^{th}$ 1-second (1024 samples) of fake vibration tensor at joint 1 in undamaged domain (0) |
| $[a_{11f\_n}]$ | $n^{th}$ 1-second (1024 samples) of fake vibration tensor at joint 1 in damaged domain (1) |

**Data Availability Statement**

Vibration data used in this study was made available (Avci et al. 2022). Some or all used models, codes, and detailed results are available from the corresponding author of this paper upon request.

**Acknowledgement**

The authors would like to thank members of the CITRS (Civil Infrastructure Technologies for Resilience and Safety) Research Initiative at the University of Central Florida. The second author would like to acknowledge the support of the National Aeronautics and Space Administration (NASA) Award No. 80NSSC20K0326.

**References**

Abdeljaber, O., and O. Avci. 2016. "Nonparametric Structural Damage Detection Algorithm for Ambient Vibration Response: Utilizing Artificial Neural Networks and Self-Organizing Maps." *Journal of Architectural Engineering*, 22 (2). https://doi.org/10.1061/(ASCE)AE.1943-5568.0000205.

Abdeljaber, O., O. Avci, N. T. Do, M. Gul, O. Celik, and F. N. Catbas. 2016a. "Quantification of Structural Damage with Self-Organizing Maps." 47–57.

Abdeljaber, O., O. Avci, M. S. Kiranyaz, B. Boashash, H. Sodano, and D. J. Inman. 2018. "1-D CNNs for structural damage detection: Verification on a structural health monitoring benchmark data." *Neurocomputing*, 275. https://doi.org/10.1016/j.neucom.2017.09.069.

Abdeljaber, O., O. Avci, S. Kiranyaz, M. Gabbouj, and D. J. Inman. 2017. "Real-time vibration-based structural damage detection using one-dimensional convolutional neural networks." *Journal of Sound and Vibration*, 388. https://doi.org/10.1016/j.jsv.2016.10.043.

Abdeljaber, O., A. Younis, O. Avci, N. Catbas, M. Gul, O. Celik, and H. Zhang. 2016b. "Dynamic Testing of a Laboratory Stadium Structure." *Geotechnical and Structural Engineering Congress 2016*, 1719–1728. Reston, VA: American Society of Civil Engineers.

An, D., J.-H. Choi, and N. H. Kim. 2013. "Prognostics 101: A tutorial for particle filter-based prognostics algorithm using Matlab." *Reliability Engineering & System Safety*, 115: 161–169. https://doi.org/10.1016/j.ress.2013.02.019.

Arjovsky, M., S. Chintala, and L. Bottou. 2017. "Wasserstein GAN."

Avci, O., O. Abdeljaber, S. Kiranyaz, M. Hussein, M. Gabbouj, and D. Inman. 2022. "A New Benchmark Problem for Structural Damage Detection: Bolt Loosening Tests on a Large-Scale Laboratory Structure." 15–22.

Avci, O., O. Abdeljaber, S. Kiranyaz, M. Hussein, M. Gabbouj, and D. J. Inman. 2021. "A review of vibration-based damage detection in civil structures: From traditional methods to Machine Learning and Deep Learning applications." *Mechanical Systems and Signal Processing*, 147: 107077. https://doi.org/10.1016/j.ymssp.2020.107077.

Avci, O., O. Abdeljaber, S. Kiranyaz, and D. Inman. 2017. "Structural Damage Detection in Real Time: Implementation of 1D Convolutional Neural Networks for SHM Applications."

Bandara, R. P., T. H. Chan, and D. P. Thambiratnam. 2014. "Structural damage detection method using frequency response functions." *Structural Health Monitoring*, 13 (4). https://doi.org/10.1177/1475921714522847.

Barrett, A. R., O. Avci, M. Setareh, and T. M. Murray. 2006. "Observations from Vibration Testing of In-Situ Structures." *Structures Congress 2006*, 1–10. Reston, VA: American Society of Civil Engineers.

Borji, A. 2018. "Pros and Cons of GAN Evaluation Measures."

Borji, A. 2021. "Pros and Cons of GAN Evaluation Measures: New Developments."

Catbas, F. N., and A. E. Aktan. 2002. "Condition and Damage Assessment: Issues and Some Promising Indices." *Journal of Structural Engineering*, 128 (8): 1026–1036. https://doi.org/10.1061/(ASCE)0733-9445(2002)128:8(1026).

Catbas, F. N., D. L. Brown, and A. E. Aktan. 2006. "Use of Modal Flexibility for Damage Detection and Condition Assessment: Case Studies and Demonstrations on Large Structures." *Journal of Structural Engineering*, 132 (11): 1699–1712. https://doi.org/10.1061/(ASCE)0733-9445(2006)132:11(1699).

Catbas, F. N., O. Celik, O. Avci, O. Abdeljaber, M. Gul, and N. T. Do. 2017. "Sensing and Monitoring for Stadium Structures: A Review of Recent Advances and a Forward Look." *Frontiers in Built Environment*, 3. https://doi.org/10.3389/fbuil.2017.00038.

Coppe, A., M. J. Pais, R. T. Haftka, and N. H. Kim. 2012. "Using a Simple Crack Growth Model in Predicting Remaining Useful Life." *Journal of Aircraft*, 49 (6): 1965–1973. https://doi.org/10.2514/1.C031808.

Costa, V., N. Lourenço, J. Correia, and P. Machado. 2019. "COEGAN." *Proceedings of the Genetic and Evolutionary Computation Conference*. New York, NY, USA: ACM.

Cury, A., and C. Crémona. 2012. "Pattern recognition of structural behaviors based on learning algorithms and symbolic data concepts." *Structural Control and Health Monitoring*, 19 (2). https://doi.org/10.1002/stc.412.

Das, S., P. Saha, and S. K. Patro. 2016. "Vibration-based damage detection techniques used for health monitoring of structures: a review." *Journal of Civil Structural Health Monitoring*, 6 (3): 477–507. https://doi.org/10.1007/s13349-016-0168-5.

Do, N. T., M. Gül, O. Abdeljaber, and O. Avci. 2018. "Novel Framework for Vibration Serviceability Assessment of Stadium Grandstands Considering Durations of Vibrations." *Journal of Structural Engineering*, 144 (2): 04017214. https://doi.org/10.1061/(ASCE)ST.1943-541X.0001941.

Drouillet, C., J. Karandikar, C. Nath, A.-C. Journeaux, M. el Mansori, and T. Kurfess. 2016. "Tool life predictions in milling using spindle power with the neural network technique." *Journal of Manufacturing Processes*, 22: 161–168. https://doi.org/10.1016/j.jmapro.2016.03.010.

Eren, L. 2017. "Bearing Fault Detection by One-Dimensional Convolutional Neural Networks." *Mathematical Problems in Engineering*, 2017. https://doi.org/10.1155/2017/8617315.

Fan, G., J. Li, H. Hao, and Y. Xin. 2021. "Data driven structural dynamic response reconstruction using segment based generative adversarial networks." *Engineering Structures*, 234. https://doi.org/10.1016/j.engstruct.2021.111970.

Farrar, C. R., and K. Worden. 2007. "An introduction to structural health monitoring." *Philosophical Transactions of the Royal Society A: Mathematical, Physical and Engineering Sciences*, 365 (1851): 303–315. https://doi.org/10.1098/rsta.2006.1928.

Ferro, R., N. Obin, and A. Roebel. 2019. "CycleGAN Voice Conversion of Spectral Envelopes using Adversarial Weights."

Gao, S., X. Wang, X. Miao, C. Su, and Y. Li. 2019. "ASM1D-GAN: An Intelligent Fault Diagnosis Method Based on Assembled 1D Convolutional Neural Network and Generative Adversarial Networks." *Journal of Signal Processing Systems*, 91 (10). https://doi.org/10.1007/s11265-019-01463-8.

Ghiasi, R., P. Torkzadeh, and M. Noori. 2016. "A machine-learning approach for structural damage detection using least square support vector machine based on a new combinational kernel function." *Structural Health Monitoring*, 15 (3): 302–316. https://doi.org/10.1177/1475921716639587.

González, M. P., and J. L. Zapico. 2008. "Seismic damage identification in buildings using neural networks and modal data." *Computers & Structures*, 86 (3–5). https://doi.org/10.1016/j.compstruc.2007.02.021.

Goodfellow, I. 2016. "NIPS 2016 Tutorial: Generative Adversarial Networks."

Goodfellow, I. J., J. Pouget-Abadie, M. Mirza, B. Xu, D. Warde-Farley, S. Ozair, A. Courville, and Y. Bengio. 2014. "Generative Adversarial Networks."

Gu, J., T. S. Yang, J. C. Ye, and D. H. Yang. 2021. "CycleGAN denoising of extreme low-dose cardiac CT using wavelet-assisted noise disentanglement." *Medical Image Analysis*, 74: 102209. https://doi.org/10.1016/j.media.2021.102209.

Guan, S., and M. Loew. 2020. "A Novel Measure to Evaluate Generative Adversarial Networks Based on Direct Analysis of Generated Images."

Gul, M., and F. N. Catbas. 2008. "Ambient Vibration Data Analysis for Structural Identification and Global Condition Assessment." *Journal of Engineering Mechanics*, 134 (8): 650–662. https://doi.org/10.1061/(ASCE)0733-9399(2008)134:8(650).

Gul, M., and F. N. Catbas. 2011. "Damage Assessment with Ambient Vibration Data Using a Novel Time Series Analysis Methodology." *Journal of Structural Engineering*, 137 (12): 1518–1526. https://doi.org/10.1061/(ASCE)ST.1943-541X.0000366.

Gulrajani, I., F. Ahmed, M. Arjovsky, V. Dumoulin, and A. Courville. 2017. "Improved Training of Wasserstein GANs."

Guo, Q., Y. Li, Y. Song, D. Wang, and W. Chen. 2020. "Intelligent Fault Diagnosis Method Based on Full 1-D Convolutional Generative Adversarial Network." *IEEE Transactions on Industrial Informatics*, 16 (3). https://doi.org/10.1109/TII.2019.2934901.

Heusel, M., H. Ramsauer, T. Unterthiner, B. Nessler, and S. Hochreiter. 2017. "GANs Trained by a Two Time-Scale Update Rule Converge to a Local Nash Equilibrium."

Hu, W., M. Li, and X. Ju. 2021. *Improved CycleGAN for Image-to-Image Translation*.

Jiang, H., C. Wan, K. Yang, Y. Ding, and S. Xue. 2021. "Continuous missing data imputation with incomplete dataset by generative adversarial networks–based unsupervised learning for long-term bridge health monitoring." *Structural Health Monitoring*. https://doi.org/10.1177/14759217211021942.

Kaneko, T., and H. Kameoka. 2017. "Parallel-Data-Free Voice Conversion Using Cycle-Consistent Adversarial Networks."

Kaneko, T., H. Kameoka, K. Tanaka, and N. Hojo. 2019. "CycleGAN-VC2: Improved CycleGAN-based Non-parallel Voice Conversion."

Krishnan Nair, K., and A. S. Kiremidjian. 2007. "Time Series Based Structural Damage Detection Algorithm Using Gaussian Mixtures Modeling." *Journal of Dynamic Systems, Measurement, and Control*, 129 (3): 285–293. https://doi.org/10.1115/1.2718241.

Kuo, P.-H., S.-T. Lin, and J. Hu. 2020. "DNAE-GAN: Noise-free acoustic signal generator by integrating autoencoder and generative adversarial network." *International Journal of Distributed Sensor Networks*, 16 (5). https://doi.org/10.1177/1550147720923529.

Kwon, D., M. R. Hodkiewicz, J. Fan, T. Shibutani, and M. G. Pecht. 2016. "IoT-Based Prognostics and Systems Health Management for Industrial Applications." *IEEE Access*, 4: 3659–3670. https://doi.org/10.1109/ACCESS.2016.2587754.

Lee, J. J., J. W. Lee, J. H. Yi, C. B. Yun, and H. Y. Jung. 2005. "Neural networks-based damage detection for bridges considering errors in baseline finite element models." *Journal of Sound and Vibration*, 280 (3–5). https://doi.org/10.1016/j.jsv.2004.01.003.

Lee, J., and S. Kim. 2007. "Structural Damage Detection in the Frequency Domain using Neural Networks." *Journal of Intelligent Material Systems and Structures*, 18 (8). https://doi.org/10.1177/1045389X06073640.

Li, W., and J. Wang. 2021. "Residual Learning of Cycle-GAN for Seismic Data Denoising." *IEEE Access*, 9: 11585–11597. https://doi.org/10.1109/ACCESS.2021.3049479.

Li, Y., H. Wang, and X. Dong. 2021. "The Denoising of Desert Seismic Data Based on Cycle-GAN With Unpaired Data Training." *IEEE Geoscience and Remote Sensing Letters*, 18 (11): 2016–2020. https://doi.org/10.1109/LGRS.2020.3011130.

Liao, L., and F. Kottig. 2014. "Review of Hybrid Prognostics Approaches for Remaining Useful Life Prediction of Engineered Systems, and an Application to Battery Life Prediction." *IEEE Transactions on Reliability*, 63 (1): 191–207. https://doi.org/10.1109/TR.2014.2299152.

Loshchilov, I., and F. Hutter. 2017. "Decoupled Weight Decay Regularization."

Luleci, F., F. N. Catbas, and O. Avci. 2021a. "Generative Adversarial Networks for Labelled Vibration Data Generation."

Luleci, F., F. N. Catbas, and O. Avci. 2021b. "Generative Adversarial Networks for Labeled Data Creation for Structural Monitoring and Damage Detection."

Luleci, F., F. N. Catbas, and O. Avci. 2022. "Generative Adversarial Networks for Data Generation in Structural Health Monitoring." *Frontiers in Built Environment*, 8. https://doi.org/10.3389/fbuil.2022.816644.

Luo, T., Y. Fan, L. Chen, G. Guo, and C. Zhou. 2020. "EEG Signal Reconstruction Using a Generative Adversarial Network With Wasserstein Distance and Temporal-Spatial-Frequency Loss." *Frontiers in Neuroinformatics*, 14. https://doi.org/10.3389/fninf.2020.00015.

Medjaher, K., D. A. Tobon-Mejia, and N. Zerhouni. 2012. "Remaining Useful Life Estimation of Critical Components With Application to Bearings." *IEEE Transactions on Reliability*, 61 (2): 292–302. https://doi.org/10.1109/TR.2012.2194175.

MITCSAIL. 2019. "Spatial Evolutionary Generative Adversarial Networks." *https://jamaltoutouh.github.io/downloads/GECCO-2019-Mustangs.pdf*.

Muhammad, Z., P. Reynolds, O. Avci, and M. Hussein. 2018. "Review of Pedestrian Load Models for Vibration Serviceability Assessment of Floor Structures." *Vibration*, 2 (1): 1–24. https://doi.org/10.3390/vibration2010001.

Ortiz, J. L., and R. Carrasco. 2016. "Model-based fault detection and diagnosis in ALMA subsystems." *Observatory Operations: Strategies, Processes, and Systems VI*, A. B. Peck, C. R. Benn, and R. L. Seaman, eds., 110. SPIE.

Pathirage, C. S. N., J. Li, L. Li, H. Hao, W. Liu, and P. Ni. 2018. "Structural damage identification based on autoencoder neural networks and deep learning." *Engineering Structures*, 172. https://doi.org/10.1016/j.engstruct.2018.05.109.

Radford, A., L. Metz, and S. Chintala. 2015. "Unsupervised Representation Learning with Deep Convolutional Generative Adversarial Networks."

Rastin, Z., G. Ghodrati Amiri, and E. Darvishan. 2021. "Unsupervised Structural Damage Detection Technique Based on a Deep Convolutional Autoencoder." *Shock and Vibration*, 2021. https://doi.org/10.1155/2021/6658575.

Sabir, R., D. Rosato, S. Hartmann, and C. Guhmann. 2021. "Signal Generation using 1d Deep Convolutional Generative Adversarial Networks for Fault Diagnosis of Electrical Machines." *2020 25th International Conference on Pattern Recognition (ICPR)*. IEEE.

Salimans, T., I. Goodfellow, W. Zaremba, V. Cheung, A. Radford, and X. Chen. 2016. "Improved Techniques for Training GANs."

Santos, A., E. Figueiredo, M. F. M. Silva, C. S. Sales, and J. C. W. A. Costa. 2016. "Machine learning algorithms for damage detection: Kernel-based approaches." *Journal of Sound and Vibration*, 363. https://doi.org/10.1016/j.jsv.2015.11.008.

Sayyad, S., S. Kumar, A. Bongale, P. Kamat, S. Patil, and K. Kotecha. 2021. "Data-Driven Remaining Useful Life Estimation for Milling Process: Sensors, Algorithms, Datasets, and Future Directions." *IEEE Access*, 9: 110255–110286. https://doi.org/10.1109/ACCESS.2021.3101284.

Shang, Z., L. Sun, Y. Xia, and W. Zhang. 2021. "Vibration-based damage detection for bridges by deep convolutional denoising autoencoder." *Structural Health Monitoring*, 20 (4). https://doi.org/10.1177/1475921720942836.

Shao, S., P. Wang, and R. Yan. 2019. "Generative adversarial networks for data augmentation in machine fault diagnosis." *Computers in Industry*, 106. https://doi.org/10.1016/j.compind.2019.01.001.

Silva, M., A. Santos, E. Figueiredo, R. Santos, C. Sales, and J. C. W. A. Costa. 2016. "A novel unsupervised approach based on a genetic algorithm for structural damage detection in bridges." *Engineering Applications of Artificial Intelligence*, 52: 168–180. https://doi.org/10.1016/j.engappai.2016.03.002.

Truong, T., and S. Yanushkevich. 2019. "Generative Adversarial Network for Radar Signal Synthesis." *2019 International Joint Conference on Neural Networks (IJCNN)*. IEEE.

Wang, T., D. Trugman, and Y. Lin. 2021. "SeismoGen: Seismic Waveform Synthesis Using GAN With Application to Seismic Data Augmentation." *Journal of Geophysical Research: Solid Earth*, 126 (4). https://doi.org/10.1029/2020JB020077.

Wang, Y., Y. Zhao, and S. Addepalli. 2020. "Remaining Useful Life Prediction using Deep Learning Approaches: A Review." *Procedia Manufacturing*, 49: 81–88. https://doi.org/10.1016/j.promfg.2020.06.015.

Wu, J., Y. Su, Y. Cheng, X. Shao, C. Deng, and C. Liu. 2018. "Multi-sensor information fusion for remaining useful life prediction of machining tools by adaptive network based fuzzy inference system." *Applied Soft Computing*, 68: 13–23. https://doi.org/10.1016/j.asoc.2018.03.043.

Wulan, N., W. Wang, P. Sun, K. Wang, Y. Xia, and H. Zhang. 2020. "Generating electrocardiogram signals by deep learning." *Neurocomputing*, 404. https://doi.org/10.1016/j.neucom.2020.04.076.

Xia, M., T. Li, L. Liu, L. Xu, and C. W. Silva. 2017. "Intelligent fault diagnosis approach with unsupervised feature learning by stacked denoising autoencoder." *IET Science, Measurement & Technology*, 11 (6): 687–695. https://doi.org/10.1049/iet-smt.2016.0423.

Xiang, Y., C. Bao, and J. Yuan. 2020. "A Weekly Supervised Speech Enhancement Strategy using Cycle-GAN." *2020 IEEE International Conference on Signal Processing, Communications and Computing (ICSPCC)*, 1–5. IEEE.

Yao, Y., J. Plested, and T. Gedeon. 2018. "A Feature Filter for EEG Using Cycle-GAN Structure." 567–576.

Yin, T., H. F. Lam, H. M. Chow, and H. P. Zhu. 2009. "Dynamic reduction-based structural damage detection of transmission tower utilizing ambient vibration data." *Engineering Structures*, 31 (9): 2009–2019. https://doi.org/10.1016/j.engstruct.2009.03.004.

Yu, Y., C. Wang, X. Gu, and J. Li. 2019. "A novel deep learning-based method for damage identification of smart building structures." *Structural Health Monitoring*, 18 (1). https://doi.org/10.1177/1475921718804132.

Zhang, C., S. R. Kuppannagari, R. Kannan, and V. K. Prasanna. 2018. "Generative Adversarial Network for Synthetic Time Series Data Generation in Smart Grids." *2018 IEEE International Conference on Communications, Control, and Computing Technologies for Smart Grids (SmartGridComm)*. IEEE.

Zhang, X., Y. Qin, C. Yuen, L. Jayasinghe, and X. Liu. 2021. "Time-Series Regeneration with Convolutional Recurrent Generative Adversarial Network for Remaining Useful Life Estimation."

Zhu, J.-Y., T. Park, P. Isola, and A. A. Efros. 2017. "Unpaired Image-to-Image Translation using Cycle-Consistent Adversarial Networks."